# Learning networks determined by the ratio of prior and data

Maomi Ueno
Graduate School of Information Systems, University of Electro-Communications
1-5-1, Chofugaoka, Chofu-shi, Tokyo, 182-8585, Japan
ueno@ai.is.uec.ac.jp

## Abstract

Recent reports have described that the equivalent sample size (ESS) in a Dirichlet prior plays an important role in learning Bayesian networks. This paper provides an asymptotic analysis of the marginal likelihood score for a Bayesian network. Results show that the ratio of the ESS and sample size determine the penalty of adding arcs in learning Bayesian networks. The number of arcs increases monotonically as the ESS increases; the number of arcs monotonically decreases as the ESS decreases. Furthermore, the marginal likelihood score provides a unified expression of various score metrics by changing prior knowledge.

## 1 INTRODUCTION

The most popular Bayesian network learning score is the marginal likelihood score (using a Dirichlet prior over model parameters), which finds the maximum a posteriori (MAP) structure, as described by Buntine (1991) and Heckerman *et al.* (1995). In addition, the Dirichlet prior is known as a distribution that ensures likelihood equivalence; this score is known as "Bayesian Dirichlet equivalence (BDe)" (Heckerman *et al.*, 1995). Given no prior knowledge, the Bayesian Dirichlet equivalence uniform (BDeu), as proposed earlier by Buntine (1991), is often used. Actually, BDeu requires an "equivalent sample size (ESS)", which is the value of a user-specified free parameter. Moreover, it has been demonstrated in recent studies that the ESS plays an important role in the resulting network structure estimate.

Steck and Jaakkola (2002) demonstrated that as the ESS asymptotically went to zero for a large sample, the deletion of an arc in a Bayesian network was favored. This result was particularly surprising because it had been believed that the likelihood, which has consistency, became dominant in the score when ESS approached zero. That study also demonstrated that when the ESS became large, the number of arcs in the structure most probably increased, which was also counterintuitive because we believed that a Bayesian prior relaxed overfitting in learning; then increasing ESS blocked the addition of extra arcs. Consequently, their findings suggested that our intuitive understanding of the Dirichlet score might have differed greatly from the correct one.

Silander, Kontkanen, and Myllymaki (2007) performed empirical experiments to find the optimum ESS of BDeu. Their results confirmed the earlier results described by Steck and Jaakkola (2002) and indicated that the solution to the network structure is highly sensitive to the chosen ESS. Nevertheless, they found no reason for the phenomenon.

Steck (2008) showed that the log-Bayes factor of dependency between two nodes using BDeu was expressible as a tradeoff between the skewness (non-uniformity) of the sample distribution and model complexity. This result was almost identical to the Akaike Information Criterion (AIC; Akaike, 1974). Additionally, Steck proposed an empirical method of optimizing the ESS to minimize the expected error measured using AIC. However, this analysis insufficiently explains BDeu's behavior with respect to the ESS. The problem with the derivation was that it did not consider effects of ESS and sample size on learning Bayesian network results.

Consequently, the marginal likelihood score mechanism has not been sufficiently explained. The main purpose of this paper is to clarify the mechanism and the role of ESS in the marginal likelihood score. First, this paper provides an asymptotic analysis of the log marginal likelihood score, which is a general form of BDe and BDeu, and its relation with other learning scores. The results indicate that the ratio of the sam-ple size and the hyperparameter determine the weight of the penalty of the number of parameters. That is, the ratio of the sample size and prior knowledge determine the Bayesian network structure.

Complementarily, the result implies that the log marginal likelihood score converges to AIC (Akaike, 1974) when the prior knowledge employs the training data and it converges to the Bayesian Information Criterion (BIC; Schwarz, 1978) when the hyperparameters are fixed at 1.0. That is, the marginal likelihood score provides a unified expression of various score metrics by changing prior knowledge.

Second, this paper provides an asymptotic analysis of the log-BDeu and explains that it can be decomposed into (1) a log-posterior that reflects the skewness (non-uniformity) of the sample distribution and (2) a penalty that blocks extra arcs from being added. Furthermore, the result shows that a tradeoff exists between the role of ESS in the log-posterior (which helps to block extra arcs) and its role in the penalty term (which helps to add extra arcs). That tradeoff might cause the BDeu score to be highly sensitive to the ESS and make it more difficult to determine an approximate ESS. In addition, this paper clarifies that the tradeoff monotonically increases the number of arcs as the ESS increases.

One argument is that learning the MAP structure is not necessarily important. One might insist that the model selection criterion should be related to the behavior of the model in some specific task such as the prediction of the next observation or classification. However, in terms of model understanding, the MAP structure, which identifies a model that is most likely to be true, is important (Chickering and Heckerman 2000).

## 2 LEARNING BAYESIAN NETWORKS

Let $\{x_1, x_2, \cdots, x_N\}$ be a set of $N$ discrete variables; each can take values in the set of states $\{1, \cdots, r_i\}$. We write $x_i = k$ when we observe that an $x_i$ is state $k$. According to the Bayesian network structure $g \in G$, the joint probabilities distribution is given as

$$p(x_1, x_2, \cdots, x_N \mid g) = \prod_{i=1}^{N} p(x_i \mid \Pi_i, g), \quad (1)$$

where $G$ is the possible set of Bayesian network structures, and $\Pi_i$ is the parent variable set of $x_i$.

Next, we introduce the problem of learning a Bayesian network. Let $\theta_{ijk}$ be a conditional probability parameter of $x_i = k$ when the $j$th instance of the parents of $x_i$ is observed (We write $\Pi_i = j$). Buntine (1991) assumed the Dirichlet prior and used an expected a posteriori(EAP) estimator as the parameter estimator $\hat{\theta}_{ijk}$:

$$\hat{\theta}_{ijk} = \frac{\alpha_{ijk} + n_{ijk}}{\alpha_{ij} + n_{ij}}, (k = 1, \cdots, r_i - 1), \quad (2)$$

where $n_{ijk}$ represents the number of samples of $x_i = k$ when $\Pi_i = j$, $n_{ij} = \sum_{k=1}^{r_i} n_{ijk}$, $\alpha_{ijk}$ denotes the hyperparameters of the Dirichlet prior distributions ($\alpha_{ijk}$ is a pseudo-sample corresponding to $n_{ijk}$), $\alpha_{ij} = \sum_{k=1}^{r_i} \alpha_{ijk}$, and $\hat{\theta}_{ijr_i} = 1 - \sum_{k=1}^{r_i-1} \hat{\theta}_{ijk}$.

The marginal likelihood is obtained as

$$p(\mathbf{X} \mid g) = \prod_{i=1}^{N} \prod_{j=1}^{q_i} \frac{\Gamma(\alpha_{ij})}{\Gamma(\alpha_{ij} + n_{ij})} \prod_{k=1}^{r_i} \frac{\Gamma(\alpha_{ijk} + n_{ijk})}{\Gamma(\alpha_{ijk})}, \quad (3)$$

where $q_i$ signifies the number of instances of $\Pi_i$, where $q_i = \prod_{x_l \in \Pi_i} r_l$, and where $\mathbf{X}$ is a dataset. The problem of learning a Bayesian network is to find the MAP structure that maximizes the score (3). We designate this score as the "marginal likelihood (ML) score".

In particular, Heckerman *et al.* (1995) presented a sufficient condition for satisfying the likelihood equivalence assumption in the form of the following constraint related to hyperparameters:

$$\alpha_{ijk} = \alpha p(x_i = k, \Pi_i = j \mid g^h). \quad (4)$$

Therein, $\alpha$ is the user-determined equivalent sample size (ESS) and $g^h$ is the hypothetical Bayesian network structure that reflects a user's prior knowledge. This metric was designated as the Bayesian Dirichlet equivalence (BDe) score metric.

As Buntine (1991) described, $\alpha_{ijk} = \frac{\alpha}{(r_i q_i)}$ is considered to be a special case of the BDe metric. Heckerman *et al.* (1995) called this special case "BDeu". Actually, $\alpha_{ijk} = \frac{\alpha}{(r_i q_i)}$ does not mean "uniform prior" but "the same value of all hyperparameters for a variable".

For fixed data and ESS, finding the MAP estimate of the structure is an NP-complete problem (Chickering, 1996). However, recently, the exact solution methods in reasonable computation time have been found if the number of variables is not prohibitively large (ex. Silander and Myllymaki, 2006).

## 3 ASYMPTOTIC ANALYSES OF THE LOG-ML SCORE

This section provides an asymptotic analysis of the ML score: a general form of BDe and BDeu.

**Theorem 1** *When $\alpha + n$ is sufficiently large, the log-ML is approximated asymptotically as*

$$\log p(\boldsymbol{X} \mid g) = \mathcal{H}(g, \alpha) - \mathcal{H}(g, \alpha, \mathbf{X}) \quad (5)$$

$$-\frac{1}{2}\sum_{i=1}^{N}\sum_{j=1}^{q_i}\sum_{k=1}^{r_i}\frac{r_i-1}{r_i}\log\left(1+\frac{n_{ijk}}{\alpha_{ijk}}\right)+O(1),$$

where $\mathcal{H}(g,\alpha), \mathcal{H}(g,\alpha,\mathbf{X})$ indicate the following empirical entropy functions.

$$\mathcal{H}(g,\alpha) = -\sum_{i=1}^{N}\sum_{j=1}^{q_i}\sum_{k=1}^{r_i}\alpha_{ijk}\log\frac{\alpha_{ijk}}{\alpha_{ij}}$$

$$\mathcal{H}(g,\alpha,\mathbf{X}) =$$
$$-\sum_{i=1}^{N}\sum_{j=1}^{q_i}\sum_{k=1}^{r_i}(\alpha_{ijk}+n_{ijk})\log\frac{(\alpha_{ijk}+n_{ijk})}{(\alpha_{ij}+n_{ij})}$$

In those equations, log represents the binary logarithm.

The proof is obtainable as the following.

**Proof 1** *From (3), log-ML is obtained as*

$$\log p(\mathbf{X}\mid g) =$$
$$\sum_{i=1}^{N}\sum_{j=1}^{q_i}\left(\sum_{k=1}^{r_i}\log\Gamma(\alpha_{ijk}+n_{ijk}) - \log\Gamma(\alpha_{ij}+n_{ij})\right)$$
$$+\sum_{i=1}^{N}\sum_{j=1}^{q_i}\left(\log\Gamma(\alpha_{ij}) - \sum_{k=1}^{r_i}\log\Gamma(\alpha_{ijk})\right).$$

*Here, we use the following Stirling series (c.f. Box and Tiao, 1992) when $a$ is sufficiently large, as*

$$\log\Gamma(a) = \frac{1}{2}\log(2\pi) + \left(a-\frac{1}{2}\right)\log a - a + \mathcal{O}\left(\frac{1}{a}\right).$$

*When $\alpha+n$ is sufficiently large, then*

$$\sum_{i=1}^{N}\sum_{j=1}^{q_i}\left(\sum_{k=1}^{r_i}\log\Gamma(\alpha_{ijk}+n_{ijk}) - \log\Gamma(\alpha_{ij}+n_{ij})\right)$$
$$= \sum_{i=1}^{N}\sum_{j=1}^{q_i}\left(\sum_{k=1}^{r_i}(\alpha_{ijk}+n_{ijk})\log(\alpha_{ijk}+n_{ijk})\right.$$
$$-(\alpha_{ij}+n_{ij})\log(\alpha_{ij}+n_{ij}) + \frac{r_i-1}{2}\log(2\pi)$$
$$\left.-\frac{1}{2}\sum_{k=1}^{r_i}\log(\alpha_{ijk}+n_{ijk}) + \frac{1}{2}\log(\alpha_{ij}+n_{ij})\right)$$
$$+\mathcal{O}(\frac{\sum_{i=1}^{N}r_iq_i}{n+\alpha})$$
$$= \sum_{i=1}^{N}\sum_{j=1}^{q_i}\sum_{k=1}^{r_i}(\alpha_{ijk}+n_{ijk})\log\frac{(\alpha_{ijk}+n_{ijk})}{(\alpha_{ij}+n_{ij})}$$
$$+\frac{1}{2}\sum_{i=1}^{N}\sum_{j=1}^{q_i}\left(\frac{r_i-1}{2}\log(2\pi) - \sum_{k=1}^{r_i}\log(\alpha_{ijk}+n_{ijk})\right.$$
$$\left.+\log(\alpha_{ij}+n_{ij})\right) + \mathcal{O}(\frac{\sum_{i=1}^{N}r_iq_i}{n+\alpha}).$$

*Similarly, we obtain*

$$\sum_{i=1}^{N}\sum_{j=1}^{q_i}\left(\log\Gamma(\alpha_{ij}) - \sum_{k=1}^{r_i}\log\Gamma(\alpha_{ijk})\right) =$$

$$-\sum_{i=1}^{N}\sum_{j=1}^{q_i}\sum_{k=1}^{r_i}\alpha_{ijk}\log\frac{\alpha_{ijk}}{\alpha_{ij}}$$
$$-\frac{1}{2}\sum_{i=1}^{N}\sum_{j=1}^{q_i}\left(\frac{r_i}{2}\log(2\pi) - \sum_{k=1}^{r_i}\log\alpha_{ijk} + \log\alpha_{ij}\right)$$
$$+\mathcal{O}(\frac{\sum_{i=1}^{N}r_iq_i}{n+\alpha}).$$

*Consequently, we obtain the following.*

$$\log p(\mathbf{X}\mid g) = \mathcal{H}(g,\alpha) - \mathcal{H}(g,\alpha,\mathbf{X}) \quad (6)$$
$$-\frac{1}{2}\sum_{i=1}^{N}\sum_{j=1}^{q_i}\left(\sum_{k=1}^{r_i}\log(\alpha_{ijk}+n_{ijk}) - \log(\alpha_{ij}+n_{ij})\right)$$
$$+\frac{1}{2}\sum_{i=1}^{N}\sum_{j=1}^{q_i}\left(\sum_{k=1}^{r_i}\log\alpha_{ijk} - \log\alpha_{ij}\right)$$
$$+\mathcal{O}(\frac{\sum_{i=1}^{N}r_iq_i}{n+\alpha})$$

*Because $\log$ function is concave, using Jensen's inequality, we obtain*

$$\frac{1}{r_i}\sum_{k=1}^{r_i}\log(\alpha_{ijk}+n_{ijk}) + \log r_i \geq \log(\alpha_{ij}+n_{ij}), \text{ and}$$

$$\frac{1}{r_i}\sum_{k=1}^{r_i}\log\alpha_{ijk} + \log r_i \geq \log\alpha_{ij}.$$

*From this, we can infer that $\sum_{k=1}^{r_i}\log(\alpha_{ijk}+n_{ijk})$ becomes dominant in the term $\sum_{i=1}^{N}\sum_{j=1}^{q_i}(\sum_{k=1}^{r_i}\log(\alpha_{ijk}+n_{ijk}) - \log(\alpha_{ij}+n_{ij}))$, and that $\sum_{k=1}^{r_i}\log\alpha_{ijk}$ becomes dominant in the term $\sum_{i=1}^{N}\sum_{j=1}^{q_i}(\sum_{k=1}^{r_i}\log\alpha_{ijk} - \log\alpha_{ij})$.*

*We approximate $\log(\alpha_{ij}+n_{ij})$ in (6) by the upper bound $\frac{1}{r_i}\sum_{k=1}^{r_i}\log(\alpha_{ijk}+n_{ijk}) + \log r_i$, and approximate $\log\alpha_{ij}$ in (6) by the upper bound $\frac{1}{r_i}\sum_{k=1}^{r_i}\log\alpha_{ijk} + \log r_i$.*

*As a consequence, we obtain*

$$\log p(\mathbf{X}\mid g) = \mathcal{H}(g,\alpha) - \mathcal{H}(g,\alpha,\mathbf{X})$$
$$-\frac{1}{2}\sum_{i=1}^{N}\sum_{j=1}^{q_i}\sum_{k=1}^{r_i}\frac{r_i-1}{r_i}\log\left(1+\frac{n_{ijk}}{\alpha_{ijk}}\right) + \mathcal{O}(1).$$

*Therefore, we obtain (5). This completes the proof.* □

Suzuki (1993) and Bouckaert (1994) also obtained asymptotic analyses of the log-ML, but the hyperparameters $\alpha_{ijk}$ did not remain in the results because they were derived using the specific hyperparameter values of 1/2 and 1.0.

Steck and Jaakkola (2002) and Steck (2008) also derived an asymptotic approximation of log-ML. Their results included the hyperparameters, but their results did not sufficiently explain the role of $\alpha_{ijk}$.

We assumed a large $\alpha + n$ instead of a large $n$ in earlier studies. This relaxed the assumption of asymptotic expansion in the previous works to retain $\alpha_{ijk}$ in the results. A unique feature of this result is that the penalty term consists of the ratio of the sample size and the prior knowledge $\log\left(1 + \frac{n_{ijk}}{\alpha_{ijk}}\right)$. This term reflects the difference between the learned structure from data and the hypothetical structure from the user's knowledge. As the two structures become equivalent, the penalty term increasingly eliminates the effect of sample size. In the opposite sense, as the two structures become different, the sample size increases the penalty term. Consequently, the user's prior knowledge more strongly reflects learning networks.

A user's prior knowledge determines learning Bayesian networks. Therefore, the traditional learning scores might be expressed by prior knowledge. Specifically, theorem 1 engenders the following relation between the ML score and the well known score metrics of AIC (Akaike, 1974) and BIC (Schwarz, 1978).

**Proposition 1** *When $\alpha_{ijk} = \frac{1}{3}n_{ijk}$ for $\forall i, \forall j, \forall k$, Then the log-ML is approximated asymptotically by AIC.*

**Proof 2** *From (5), when $\alpha_{ijk} = \frac{1}{3}n_{ijk}$, we obtain*

$$\log p(\boldsymbol{X} \mid g) = \mathcal{H}(g, \alpha) + \frac{4}{3}l(\hat{\theta} \mid \mathbf{X}) \\ - \frac{1}{2}\sum_{i=1}^{N}q_i(r_i - 1)\log 4,$$

*where $l(\hat{\theta} \mid \mathbf{X})$ is the log likelihood.*

*When $\alpha_{ijk} = \frac{1}{3}n_{ijk}$, we obtain $\mathcal{H}(g, \alpha) = -\frac{1}{3}l(\hat{\theta} \mid \mathbf{X})$. Accordingly, we obtain $\log p(\boldsymbol{X} \mid g) = -AIC$, where $AIC = -l(\hat{\theta} \mid \mathbf{X}) + \sum_{i=1}^{N}q_i(r_i - 1)$. This completes the proof.* □

In fact, AIC can be interpreted as an approximation of the test error in cross-validation by the training data (Akaike 1974). Proposition 1 means exactly this. If we have some data $\mathbf{X_{prior}}$ for learning, then $\alpha_{ijk}$ in the ML score can be estimated as the number of samples of $x_i = k$ when $\Pi_i = j$ from $\mathbf{X_{prior}}$. Consequently, if we obtain different new data $\mathbf{X}$ with the same size as $\mathbf{X_{prior}}$, then $n_{ijk}$ in the ML score can be estimated from $\mathbf{X}$. In this case, the penalty term in (5) becomes smallest when the hypothetical structure is the true one because $\alpha_{ijk}$ approaches $n_{ijk}$, which helps the score to select the true structure because the penalty increases as the structure increasingly differs from the true one. However, Proposition 1 engenders AIC using the training data $\mathbf{X}$ again instead of the prior data $\mathbf{X_{prior}}$. Unfortunately, that often causes an overfitting problem because the pseudo-augmented data as a prior do not change the likelihood in (5). Nevertheless, the penalty term is constantly small even when the hypothetical structure is wrong. Adding extra arcs increases the likelihood. Therefore, the score tends to add extra arcs. This result agrees with the well-known AIC overfitting characteristics.

**Proposition 2** *For $\forall i, \forall j, \forall k$, $\alpha_{ijk} = 1$(uniform prior), when $n$ is sufficiently large, the log-ML is approximated asymptotically by BIC.*

**Proof 3** *When $\alpha_{ijk} = 1$, we obtain $\mathcal{H}(g, \alpha) = \frac{1}{2}\sum_{i=1}^{N}q_i r_i \log r_i^2$. From (5) when $\alpha_{ijk} = 1$, we obtain*

$$\log p(\boldsymbol{X} \mid g) = -\mathcal{H}(g, \alpha, \mathbf{X}) \\ - \frac{1}{2}\sum_{i=1}^{N}q_i(r_i - 1)\log(1 - r_i^2 + n_{ijk}).$$

*When $n \to \infty$,*

$$-\mathcal{H}(g, \alpha, \mathbf{X}) \to l(\hat{\theta} \mid \mathbf{X}), and$$

$$\frac{1}{2}\sum_{i=1}^{N}q_i(r_i - 1)\log(1 - r_i^2 + n_{ijk}) \\ \to \frac{1}{2}\sum_{i=1}^{N}q_i(r_i - 1)\log n_{ijk} \\ \leq \frac{1}{2}\sum_{i=1}^{N}q_i(r_i - 1)\log n.$$

*Consequently, when $\alpha_{ijk} = 1$, then $\log p(\boldsymbol{X} \mid g) \to BIC$, where $BIC = l(\hat{\theta} \mid \mathbf{X}) - \frac{1}{2}\sum_{i=1}^{N}q_i(r_i - 1)\log n$. This completes the proof.* □

The log-ML with $\alpha_{ijk} = 1$ behaves identically to BIC when $n$ is sufficiently large. In contrast to AIC, the ML score with $\alpha_{ijk} = 1$ has consistency because BIC has consistency. In addition, the ML with $\alpha_{ijk} = 1$ is equivalent to the score metric in K2 (Cooper and Herskovits, 1992). Moreover, Bouckaert (1994) derived an MDL-based criterion that was equivalent to BIC from the log-ML with a fixed hyperparameter $\alpha_{ijk} = 1$, which agrees with our derivation. It is noteworthy that the ML with $\alpha_{ijk} = 1$ does not satisfy the likelihood equivalence, although the BIC does.

## 4 ASYMPTOTIC ANALYSES OF BDeu

For cases in which we have no prior knowledge, the Bayesian Dirichlet equivalence uniform (BDeu) is often used in practice. Actually, Heckerman *et al.* (1995) reported, as a result of their comparative analysis between BDeu and BDe, that BDeu was better than BDe unless the user's beliefs are close to the true model.

Actually, BDeu requires an "equivalent sample size (ESS)", which is the value of a free parameter to be specified by the user. Recent reports have described that the equivalent sample size (ESS) in BDeu plays an important role in learning Bayesian networks.

Steck and Jaakkola (2002) demonstrated that as the ESS asymptotically went to zero for a large sample, the deletion of an arc in a Bayesian network was favored. They also showed that when ESS became large, the number of arcs in the structure usually increased. However, they found no readily apparent reason for the phenomenon. Our assumption, that a large $\alpha + n$, expresses the conditions—a small $\alpha$ and a large $n$, and a large $\alpha$ and a small $n$—in which Steck and Jaakkola (2002) found the phenomenon. This section presents an explanation showing that Theorem 1 can explain these phenomena.

Theorem 1 readily engenders the following corollary.

**Corollary 1** *When $\alpha + n$ is sufficiently large, log-BDeu converges to*

$$\log p(\boldsymbol{X} \mid g) = \alpha \sum_{i=1}^{N} \log r_i - \mathcal{H}(g, \alpha, \mathbf{X}) \quad (7)$$
$$- \frac{1}{2} \sum_{i=1}^{N} \sum_{j=1}^{q_i} \sum_{k=1}^{r_i} \frac{r_i - 1}{r_i} \log\left(1 + \frac{r_i q_i n_{ijk}}{\alpha}\right).$$

From this, log-BDeu can also be divided into two parts: (1) log-posterior term $-\mathcal{H}(g, \alpha, \mathbf{X})$ and (2) penalty term $\frac{1}{2} \sum_{i=1}^{N} \sum_{j=1}^{q_i} \sum_{k=1}^{r_i} \frac{r_i-1}{r_i} \log\left(1 + \frac{(r_i q_i n_{ijk})}{\alpha}\right)$.

This well known model selection formula is generally interpreted as (1) reflecting the fit to the data and (2) signifying the penalty that blocks extra arcs from being added.

The main difference from results of Steck (2008) is that we demonstrate that the ESS and sample size have a strong impact on learning. Our expression shows that the sample size affects the log-posterior, strongly reflecting the features of the empirical distribution. In turn, this increases the strength of the penalty, which is necessary to block arcs from being added. However, from (7), ESS $\alpha$ in the log-posterior increases the empirical entropy (uniformity), $\mathcal{H}(g, \alpha, \mathbf{X})$, and helps to block the skewness of the sample distribution from increasing. Furthermore, the penalty term $\frac{1}{2} \sum_{i=1}^{N} \sum_{j=1}^{q_i} \sum_{k=1}^{r_i} \frac{r_i-1}{r_i} \log\left(1 + \frac{(r_i q_i n_{ijk})}{\alpha}\right)$ decreases monotonically as $\alpha$ increases. That is, $\alpha$ in the penalty term helps to add arcs. These results mean that the ESS in BDeu plays completely contrary roles from that of the sample size. Additionally, our results indicate a tradeoff between the role of the ESS in the log-posterior, which helps to block extra arcs from being added, and the role of the ESS in the penalty-term (which helps extra arcs to be added). Moreover, the important thing is that this tradeoff of $\alpha$ clearly indicates why the BDe(u) score is highly sensitive to ESS.

In addition, the following proposition is derived.

**Proposition 3** *When $\alpha$ approaches infinity, the structure learned using BDeu approaches a complete graph.*

**Proof 4** *When $\alpha \to \infty$, the penalty term $\frac{1}{2} \sum_{i=1}^{N} \sum_{j=1}^{q_i} \left( \sum_{k=1}^{r_i} \frac{r_i - 1}{r_i} \log(1 + \frac{(r_i q_i n_{ijk})}{\alpha}) \right)$ in (7) converges to 0. Consequently, log-BDeu converges to*

$$\log p(\boldsymbol{X} \mid g) = -\mathcal{H}(g, \alpha, \mathbf{X}) + \alpha \sum_{i=1}^{N} \log r_i + \mathcal{O}(1) \quad (8)$$

*Term $\alpha \sum_{i=1}^{N} \log r_i$ is constant for the number of arcs and term $-\mathcal{H}(g, \alpha, \mathbf{X})$ increases monotonically as the number of arcs $q_i$ increases. This completes the proof.* □

As previously described, there is a tradeoff in ESS $\alpha$ between the log-posterior that is blocking arcs from being added and the penalty term that is trying to add arcs. However, the penalty term converges to 0 when the ESS becomes sufficiently large. Consequently, only the log-posterior term works. Therefore, the number of arcs increases monotonically when $\alpha$ becomes sufficiently large. This result was demonstrated by Silander, Kontkanen, and Myllymaki (2007) using actual data.

On the other hand, Steck and Jaakkola (2002) suggested that as the ESS asymptotically went to zero for a large sample, the deletion of an arc in a Bayesian network was favored and an empty graph was thereby obtained. However, this problem is more complicated than the case of infinite ESS because the tradeoff remains even when the ESS becomes extremely small. Accordingly, the addition or deletion of an arc depends on the magnitude of the sample size. Suzuki (2006) derived the sufficient condition of strong consistency, $\log \log n < c < n$, when the score for learning Bayesian networks is described as $l(\hat{\theta} \mid \mathbf{X}) - c \times \frac{k}{2}$ where $k$ is the number of parameters. From (7), when the ESS approaches to zero, the asymptotically sufficient condition of strong consistency for BDeu is $\frac{r_i q_i n}{\exp n - 1} < \alpha < \frac{r_i q_i n}{\log n - 1}$ using $n_{ijk} \leq n$. Therefore, strictly considered, the consistency does not hold mathematically when the ESS approaches zero. However, if $n \to \infty$, then the condition for the consistency becomes $\alpha \in (0, \infty)$. Consequently, the consistency usually holds even when the ESS is an extremely small value. Even if the sample size is fixed, then the ESS value must necessarily be extremely small to obtain an empty graph. For example, Silander, Kontkanen, and Myllymaki (2007) observed an empty graph in learning networks from actual data when the ESS was $2e-20$ for a 1,484 sample size.

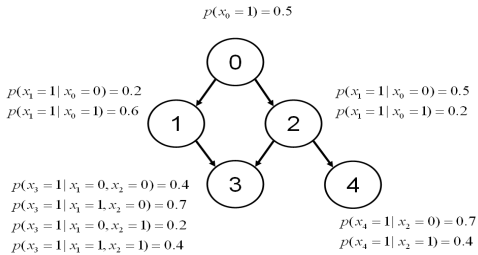

Figure 1: Structure 1 (skewed distribution)

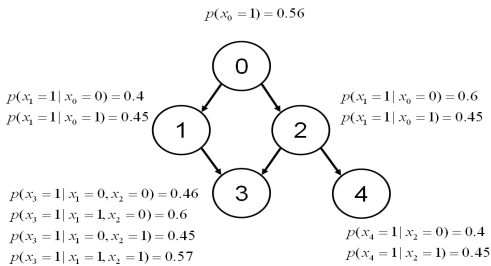

Figure 2: Structure 2 (non-skewed distribution)

## 5  NUMERICAL EXAMPLES

This section presents some results obtained from simulation experiments that confirmed the derived properties discussed in this paper. We used small network structures with binary variables in Figs. 1 and 2 for this study. Figure 1 presents a structure in which the conditional probabilities differ largely because of the parent variable states (skewed conditional distribution). Figure 2 displays a structure in which the conditional probabilities are almost identical for the parent variable states (non-skewed conditional distribution). Procedures used for the simulation experiments are described below.

1. We generated 100, 500, 1,000, 5,000, and 10,000 samples from the two figures.

2. Using AIC, BIC (MDL), and the marginal likelihood (ML) ($\alpha_{ijk} = 1/3n_{ijk}$ and $\alpha_{ijk} = 1$), Bayesian network structures were estimated, respectively, based on 100, 500, 1,000, 5,000 and 10,000 samples. Each MAP structure was searched from all possible structures.

3. The times the estimated structure was the true structure were counted by repeating Procedure 2 100 times.

Table 1 presents the results. Column "+" presents the total number of extra arcs for the estimated structures; and column "-" has the total number of missing arcs for the estimated structures. The maximum quantities of "+" and "-" were both 500. Column "O" shows the number of correct-structure estimates in 100 trials.

Table 1 shows that the learning for structure 1 was better than that of structure 2. The reason is that the skewed conditional distributions of the variables in structure 1 help the model selection criteria to detect the dependences among variables correctly. In contrast, it is difficult to detect the dependences among variables correctly using non-skewed conditional distributions of the variables as structure 2.

The results reveal that learning with AIC tends to overfit the data. The results also show that AIC and ML with $\alpha_{ijk} = 1/3n_{ijk}$ perform very similarly, as Proposition 1 shows.

Table 1 also shows that ML with $\alpha_{ijk} = 1$ behaves similarly to BIC although they are not completely equivalent. These results confirm Proposition 2.

Moreover, the results suggest that, although BIC (MDL), performs better than AIC for structure 1, it behaves similarly to AIC for structure 2. The reason is explainable using derivation (5), which can be approximated by BIC. The penalty term $\sum_{i=1}^{N} \sum_{j=1}^{q_i} \sum_{k=1}^{r_i} \frac{r_i-1}{r_i} \log\left(1 + \frac{n_{ijk}}{\alpha_{ijk}}\right)$ in (5), when $\alpha_{ijk}$ is constant, approaches constant as the conditional distribution $n_{ijk}$ approaches uniform. That is to say, ML with $\alpha_{ijk} = 1$ and BIC(MDL) behaves similarly to AIC when the conditional distributions of variables are non-skewed because the penalty term approaches the constant.

Next, we investigate learning with BDeu by changing ESS $\alpha$. Table 2 presents learning with BDeu using the same procedure as that used in previous experiments by changing the ESS value, $\alpha$ ($\alpha = 10^{-6}, 0.01, 0.1, 1, 10, 100, 10^6$).

Column $\alpha^*$ indicates the optimum $\alpha$ that provides the best performance by changing the value from 1 to 100. The results reveal that the optimum values of $\alpha$ differ greatly because of the structures and sample size. The results reflect that the optimum value of $\alpha$ is distributed around a small value when the conditional distributions are skewed, as presented in Fig. 1. In contrast, this result shows that the optimum value of $\alpha$ is distributed around a large value when the conditional distributions are almost uniform, as presented in Fig. 2.

Table 1: Comparison of AIC, BIC (MDL), and ML($\alpha_{ijk} = 1/3n_{ijk}$ and $\alpha_{ijk} = 1$)

| Structure: 1 | AIC | | | BIC(MDL) | | | ML($\alpha=1/3n_{ijk}$) | | | ML($\alpha=1$) | | |
|---|---|---|---|---|---|---|---|---|---|---|---|---|
| n | ◯ | + | - | ◯ | + | - | ◯ | + | - | ◯ | + | - |
| 100 | 0 | 369 | 23 | 0 | 369 | 23 | 27 | 91 | 56 | 17 | 77 | 83 |
| 500 | 17 | 171 | 0 | 83 | 23 | 0 | 18 | 126 | 0 | 88 | 14 | 0 |
| 1000 | 16 | 151 | 0 | 88 | 12 | 0 | 16 | 153 | 0 | 92 | 8 | 0 |
| 5000 | 26 | 133 | 0 | 98 | 2 | 0 | 22 | 148 | 0 | 99 | 1 | 0 |
| 10000 | 27 | 118 | 0 | 99 | 1 | 0 | 18 | 145 | 0 | 100 | 0 | 0 |
| Structure: 2 | AIC | | | BIC(MDL) | | | ML($\alpha=1/3n_{ijk}$) | | | ML($\alpha=1$) | | |
| n | ◯ | + | - | ◯ | + | - | ◯ | + | - | ◯ | + | - |
| 100 | 0 | 249 | 246 | 0 | 249 | 246 | 0 | 141 | 291 | 0 | 71 | 356 |
| 500 | 2 | 118 | 195 | 0 | 30 | 267 | 3 | 189 | 168 | 0 | 17 | 295 |
| 1000 | 1 | 120 | 161 | 0 | 16 | 229 | 3 | 170 | 131 | 0 | 2 | 290 |
| 5000 | 21 | 95 | 50 | 4 | 2 | 113 | 16 | 157 | 47 | 0 | 0 | 160 |
| 10000 | 23 | 95 | 31 | 10 | 0 | 90 | 19 | 137 | 27 | 5 | 0 | 97 |

Table 2: Learning with BDeu by changing ESS $\alpha$

| Structure 1 | | BDeu($\alpha=10^{-6}$) | | | BDeu($\alpha=0.01$) | | | BDeu($\alpha=0.1$) | | | BDeu($\alpha=1$) | | |
|---|---|---|---|---|---|---|---|---|---|---|---|---|---|
| n | $\alpha^*$ | ◯ | + | - | ◯ | + | - | ◯ | + | - | ◯ | + | - |
| 100 | 27 | 0 | 0 | 493 | 0 | 1 | 358 | 0 | 3 | 259 | 2 | 11 | 178 |
| 500 | 4 | 0 | 0 | 142 | 18 | 0 | 83 | 60 | 0 | 40 | 91 | 2 | 7 |
| 1000 | 1 | 5 | 0 | 95 | 98 | 0 | 2 | 100 | 0 | 0 | 100 | 0 | 0 |
| 5000 | 1 | 100 | 0 | 0 | 100 | 0 | 0 | 100 | 0 | 0 | 100 | 0 | 0 |
| 10000 | 1 | 100 | 0 | 0 | 100 | 0 | 0 | 100 | 0 | 0 | 100 | 0 | 0 |
| Structure 1 | | BDeu($\alpha=10$) | | | BDeu($\alpha=100$) | | | BDeu($\alpha=10^6$) | | | | | |
| n | $\alpha^*$ | ◯ | + | - | ◯ | + | - | ◯ | + | - | | | |
| 100 | 27 | 19 | 52 | 90 | 15 | 163 | 33 | 1 | 332 | 13 | | | |
| 500 | 4 | 89 | 11 | 0 | 42 | 77 | 0 | 0 | 460 | 0 | | | |
| 1000 | 1 | 93 | 7 | 0 | 47 | 66 | 0 | 0 | 489 | 0 | | | |
| 5000 | 1 | 100 | 0 | 0 | 90 | 10 | 0 | 0 | 500 | 0 | | | |
| 10000 | 1 | 100 | 0 | 0 | 97 | 3 | 0 | 0 | 500 | 0 | | | |
| Structure 2 | | BDeu($\alpha=10^{-6}$) | | | BDeu($\alpha=0.01$) | | | BDeu($\alpha=0.1$) | | | BDeu($\alpha=1$) | | |
| n | $\alpha^*$ | ◯ | + | - | ◯ | + | - | ◯ | + | - | ◯ | + | - |
| 100 | 42 | 0 | 0 | 500 | 0 | 0 | 499 | 0 | 0 | 489 | 0 | 8 | 458 |
| 500 | 60 | 0 | 0 | 499 | 0 | 0 | 462 | 0 | 0 | 426 | 24 | 4 | 76 |
| 1000 | 49 | 0 | 0 | 486 | 0 | 0 | 377 | 0 | 0 | 337 | 0 | 0 | 359 |
| 5000 | 92 | 0 | 0 | 300 | 0 | 0 | 251 | 0 | 0 | 207 | 0 | 2 | 290 |
| 10000 | 95 | 0 | 0 | 272 | 0 | 0 | 155 | 0 | 0 | 129 | 0 | 0 | 109 |
| Structure 2 | | BDeu($\alpha=10$) | | | BDeu($\alpha=100$) | | | BDeu($\alpha=10^6$) | | | | | |
| n | $\alpha^*$ | ◯ | + | - | ◯ | + | - | ◯ | + | - | | | |
| 100 | 42 | 0 | 51 | 362 | 0 | 158 | 281 | 0 | 219 | 238 | | | |
| 500 | 60 | 0 | 15 | 284 | 2 | 50 | 197 | 0 | 279 | 143 | | | |
| 1000 | 49 | 0 | 9 | 248 | 0 | 89 | 222 | 1 | 291 | 112 | | | |
| 5000 | 92 | 2 | 0 | 123 | 11 | 10 | 94 | 0 | 385 | 24 | | | |
| 10000 | 95 | 4 | 1 | 98 | 17 | 4 | 83 | 0 | 446 | 13 | | | |

Steck (2008) suggested, using his asymptotic derivation, that the optimal ESS becomes small when the conditional distributions of the variables are skewed. That result agrees with the results obtained from this experiment.

Using our derivation (7) about BDeu, it can be interpreted as follows: BDeu might suffer overfitting when the conditional distributions are skewed because the log-posterior in (7) sensitively detects dependences among variables. To prevent overfitting, ESS should become small and function not by adding extra arcs. On the contrary, BDeu is difficult to detect the dependences among variables when the conditional distributions are not skewed. To prevent underfitting, ESS should become large and work to add correct arcs.

Furthermore, we confirmed that the number of extra arcs increased in BDeu when the ESS increased and that the number of missing arcs increased when the ESS decreased. The estimated network structures with $\alpha = 10^6$ include many complete graphs for structures 1 and 2. In contrast, the estimated network structures with $\alpha = 10^{-6}$ include many empty graphs in structure 1 and 2. The structure 1 includes fewer empty graphs than the structure 2 because the skewed condtional distributions help to add extra arcs. Additionally, the results confirm that learning recovers as the sample size increases, even with $\alpha = 10^{-6}$, because the consistency of BDeu holds. These results confirm Proposition 3.

Table 2 also shows that BDeu with $\alpha = 1.0$ performs better than the other ESS values for both structures

1 and 2 when the sample size is small. In the case of sparse data, if we set a large ESS value, then the penalty term in (7) decreases and might suffer overfitting. To prevent this problem, we should set the ESS value such that it affects parameter estimation to the least degree possible. Therefore, from the penalty term in (7), the BDeu with $\alpha = 1$ corresponds to the smallest positive assignment of the hyperparameters, which allows the data to reflect the estimated parameters to the greatest degree possible. Additionally, it is known that the variances of the Dirichlet distribution decrease with the sum of the hyperparameters (Castillo, Hadi, and Solares, 1997). This decrease of variance also suggests that BDeu with $\alpha = 1$ is the best method to mitigate the influence of ESS for parameter estimation. In fact, ESS values smaller than 1.0 are allowed when the sample is sufficiently large. Consequently, ESS value 1.0 is recommended for learning Bayesian networks especially using sparse data.

# 6 CONCLUSIONS

We provided an asymptotic analysis of the marginal likelihood score and its relation with other learning scores. Results show that the ratio of the ESS and sample size determined the penalty of adding arcs in learning Bayesian networks. Furthermore, the result shows that the log-marginal likelihood score provided a unified expression of various score metrics by changing prior knowledge. We then presented an asymptotic analysis of log-BDeu and demonstrated that it can be decomposed into (1) a log-posterior that reflected the skewness (non-uniformity) of the sample distribution and (2) a penalty that blocked extra arcs from being added. Additionally, we showed that a tradeoff existed between the role of ESS in the log-posterior (which helps to block extra arcs) and its role in the penalty term (which helps to add extra arcs). That tradeoff might cause the BDeu score to be highly sensitive to the ESS and make it more difficult to determine an approximate ESS. Additionally, the result shows that this tradeoff increased the number of arcs monotonically with the increase of the ESS value.